%% file: ms.tex
\documentclass[sigconf]{acmart}

\usepackage{amsfonts,amsmath,amssymb}

\usepackage[utf8]{inputenc}
\usepackage{graphicx}
\usepackage{verbatim} 

\usepackage{booktabs} 
\usepackage{lineno}
\usepackage{caption}
\usepackage{booktabs}
\usepackage{algorithm}
\usepackage[noend]{algpseudocode}
\usepackage{longtable}
\usepackage{multicol}
\usepackage{rotating}
\usepackage{placeins}
 \usepackage{multirow}
\usepackage{hyperref}

\usepackage{xspace}
\usepackage{graphicx}
\usepackage{paralist}
\usepackage{url}
\usepackage{matharticle}
\usepackage{subfig}

\usepackage{xcolor}
\usepackage{colortbl}
\definecolor{GAINSBORO}{HTML}{DCDCDC}
\definecolor{LIGHTYELLOW}{HTML}{FFFFE0}
\definecolor{MISTYROSE}{HTML}{FFE4E1}
\definecolor{ALICEBLUE}{HTML}{F0F8FF}

\acmDOI{10.1145/nnnnnnn.nnnnnnn}

\acmISBN{978-x-xxxx-xxxx-x/YY/MM}

\acmConference[GECCO '21]{the Genetic and Evolutionary Computation Conference 2021}{July 10--14, 2021}{Lille, France}
\acmYear{2021}
\copyrightyear{2021}

\acmPrice{15.00}

\usepackage{graphicx}        
\usepackage{multicol}        
\usepackage[bottom]{footmisc}

\usepackage{newtxmath}       
\usepackage{xspace}
\usepackage{graphicx}
\usepackage{paralist}
\usepackage{url}
\usepackage{subfig}

\usepackage{algorithm}
\usepackage{algorithmicx}
\usepackage{algpseudocode}

\definecolor{GAINSBORO}{HTML}{DCDCDC}
\definecolor{LIGHTYELLOW}{HTML}{FFFFE0}
\definecolor{MISTYROSE}{HTML}{FFE4E1}
\definecolor{ALICEBLUE}{HTML}{F0F8FF}
\usepackage{pifont}

\newcommand{\ring}{\texttt{Lipi-Ring}\xspace}

\newcommand{\ringtime}{\texttt{Ring$^{t}$(1)}\xspace}
\newcommand{\ringone}{\texttt{Ring(1)}\xspace}
\newcommand{\ringfive}{\texttt{Ring(2)}\xspace}
\newcommand{\Lipi}{\texttt{Lipizzaner}\xspace}

\algblock{ParFor}{EndParFor}
\algnewcommand\algorithmicparfor{\textbf{parfor}}
\algnewcommand\algorithmicpardo{\textbf{do}}
\algnewcommand\algorithmicendparfor{\textbf{end\ parfor}}
\algrenewtext{ParFor}[1]{\algorithmicparfor\ #1\ \algorithmicpardo}
\algrenewtext{EndParFor}{\algorithmicendparfor}

\algrenewcommand\alglinenumber[1]{\tiny #1:}

\newcommand{\nameAlgLipiTraining}{\texttt{CoevGANsTraining}\xspace}

\newcommand{\alg}{\texttt{Algorithm}\xspace}

\begin{document}
\title[Signal Propagation in a Gradient-Based and Evolutionary Learning System]{Signal Propagation in a\\Gradient-Based and Evolutionary Learning System}

%

\author{Jamal Toutouh}
\affiliation{%
  \institution{Massachusetts Institute of Technology, CSAIL, USA}
  \institution{University of Malaga, LCC, Spain}
}
\email{toutouh@mit.edu, jamal@lcc.uma.es}

\author{Una-May O'Reilly}
\affiliation{%
  \institution{Massachusetts Institute of Technology, CSAIL}
}\email{unamay@csail.mit.edu}

\renewcommand{\shortauthors}{J. Toutouh and U. O'Reilly}

\begin{abstract}
\input{abstract}
\end{abstract}

%
%



\maketitle

\input{introduction}
\input{background}
\input{ring-gan}
\input{experimental-setup}
\input{experimental-analysis}

\input{conclusions}

\bibliographystyle{ACM-Reference-Format}
\bibliography{bibliography} 

\end{document}

%% file: abstract.tex
Generative adversarial networks (GANs) exhibit training pathologies that can lead to convergence-related degenerative behaviors, whereas spatially-distributed, coevolutionary algorithms (CEAs) for GAN training, e.g. \Lipi, are empirically robust to them. The robustness arises from diversity that occurs by training populations of generators and discriminators in each cell of  a toroidal grid.  Communication, where signals in the form of parameters of the best GAN in a cell propagate in four directions: North, South, West and East, also plays a role, by communicating adaptations that are both new and fit. We propose \ring, a distributed CEA like \Lipi, except that it uses a different spatial topology, i.e. a ring. Our central question is whether the different directionality of signal propagation (effectively migration to one or more neighbors  on each side of a cell) meets or exceeds the performance quality and training efficiency of \Lipi Experimental analysis on different datasets (i.e, MNIST, CelebA, and COVID-19 chest X-ray images) shows that there are no significant differences between the performances of the trained generative models by both methods. However, \ring significantly reduces the computational time (14.2\%$\dots$41.2\%). Thus, \ring offers an alternative to \Lipi when the computational cost of training matters.

%% file: introduction.tex
\section{Introduction}
\label{sec:introduction}

Generative modeling aims to learn a function that describes a latent distribution of a dataset. 
In a popular paradigm, a generative adversarial network (GAN) combines two deep neural networks (DNN), a generator and a discriminator, that engage in adversarial learning to optimize their weights~\cite{goodfellow2014generative}. 
The generator is trained to produce \textit{fake} samples (given an input from a random space) to fool the discriminator. The discriminator learns to discern the \textit{real} samples from the ones produced by the generator. 
This training is formulated as a minmax optimization problem through the
definitions of discriminator and generator loss, which converges when an optimal generator approximates the true
distribution so well that the discriminator only provides a random label for any sample.  

Early GAN training methods led to vanishing gradients~\cite{arjovsky2017towards} and mode collapse~\cite{arora2017gans} among other pathologies. They arose from the inherent adversarial setup of the paradigm. Several methods have been proposed to improve GAN models and have produced strong results~\cite{jolicoeur2018relativistic,mao2017least,zhao2016energy,arjovsky2017wasserstein}. 
However, GANs remain notoriously hard to train~\cite{salimans2016improved,gulrajani2017improved}. 

Using forms of evolutionary computation (EC) for GAN training has led to promising approaches. 
Evolutionary (EAs) and coevolutionary algorithms (CEAs) for weight training or spatial systems~\cite{schmiedlechner2018lipizzaner,costa2019coevolution,schmiedlechner2018towards,toutouh2019,wang2018evolutionary,toutouh2020}.  Deep neuroevolution offers concurrent architecture and weight search~\cite{costa2019coevolution}. Pareto approximations also have been proposed to define multi-objective GAN training~\cite{garciarena2018evolved}.  This variety of approaches use different ways to guide populations of networks towards convergence, while maintaining diversity and discarding \textit{problematic} (weak) individuals. 
They have been empirically demonstrated to be comparable and better than baseline GAN training methods. 

\sloppy
In this work, we focus on spatially-distributed, competitive CEAs (Comp-CEAs), such as \Lipi~\cite{schmiedlechner2018lipizzaner}. 
In these methods, the members of two populations (generators and discriminators) are placed in the cells of a toroidal geometric space (i.e., each cell contains a generator-discriminator pair). Each cell has neighbors from which it copies their pairs of generator and discriminator.   This creates sub-populations of GANs in each cell. Gradient-based training is done pairwise between the best pairing within a sub-population. Each training iteration (epoch), selection, mutation, and replacement are applied and then, the best generator-discriminator pair is updated and the remainder of the sub-populations re-copied from the neighborhood. This update and refresh effectively propagate signals along the paths of neighbors that run across . Thus, the neighborhood defines the directionality, space of signal propagation a.k.a. migration~\cite{schmiedlechner2018towards}.
Communicating adaptations that are both new and fit promotes diversity during this training process.
This diversity has been shown to disrupt premature convergence in the form of an oscillation or moving the search away from an undesired equilibria, improving the robustness to the main GAN pathologies~\cite{schmiedlechner2018lipizzaner,toutouh2020_ppsn}. 

In this work, we want to evaluate the impact of the spatial topology used by this kind of method, changing the two-dimensional toroidal grid used by \Lipi into a ring topology. 
Thus, we propose \ring. 
\ring raises central questions about the impact of the new directionality of the signal propagation given a ring.  How are  performance quality, population diversity, and computational cost impacted?  

Thus, in this paper, we pose the following research questions: 
\textbf{RQ1:} \textit{What is the effect on the generative model trained when changing the directionality of the signal propagation from four directions to two?}
\textbf{RQ2:} \textit{When the signal is propagated to only two directions, what is the impact of performing migration in one or more neighbors?}
In terms of population diversity, 
\textbf{RQ3:} \textit{How does diversity change over time in a ring topology?} 
\textit{How does diversity compare with a ring topology with different neighborhood radius?} 
\textit{How does diversity compare between ring topology and 2D grid methods, where both methods have the same sub-population size and neighborhood, but different signal directionality?}

The main contributions of this paper are: 
\textit{i)} \ring, a new distributed Comp-CEA GAN training method based on ring topology that demonstrates markedly decreased computational cost over a 2D topology, without negatively impacting training accuracy. 
\textit{ii)} an open source software implementation of \ring\footnote{\ring source code - \texttt{https://github.com/xxxxxxxxx}}, and
\textit{iii)} evaluating different variations of \ring, comparing them to \Lipi, in a set of benchmarks based on MNIST, CelebA, and COVID-19 X-Ray chest images datasets.

The rest of the paper is organized as
follows. Section~\ref{sec:related-work} presents related
work. Section~\ref{sec:ring-topology} describes the \ring method. The experimental
setup and the results are in sections~\ref{sec:experimental-setup}
and~\ref{sec:experimental-analysis}. 
Finally, conclusions are drawn and future work is outlined in Section~\ref{sec:conclusions}.

%% file: background.tex
\section{Background}
\label{sec:related-work}

This section introduces the main concepts in GANs training and
summarizes relevant studies related to this research. 

\vspace{-0.2cm}
\subsection{General GAN training}
\label{sec:gan-training}

GANs train two DNN, a generator ($G_g$) and a discriminator ($D_d$), in an adversarial setup. 
Here, $G_g$ and $D_d$ are functions parametrized by $g$ and $d$, 
where $g \in \calU$ and $d \in \calV$ with $\calU, \calV \subseteq \R^{p}$
represent the respective parameters space of both functions. 

Let $G_*$ be the target unknown distribution to which we would like to fit our generative model~\cite{arora2017generalization}. 
The generator $G_g$ receives a variable from a latent space $z\sim P_{z}(z)$ and creates a sample from data space $x = G_{g}(z)$. 
The discriminator $D_d$ assigns a probability \hbox{$p = D_{d}(x) \in [0, 1]$} that represents the likelihood that the $x$ belongs to the real training dataset, i.e., $G_*$ by applying a \textit{measuring function} $\phi:[0,1] \to \mathbb{R}$. 
The $P_{z}(z)$ is a prior on $z$ (a uniform $[-1, 1]$ distribution is typically chosen). The goal of GAN training is to find $d$ and $g$ parameters to optimize the objective function $\mathcal{L}(g,d)$.

\begin{equation}
\vspace{-0.3cm}
\label{eq:gan-def}
\min_{g\in \mathbb{G}}\max_{d \in \mathbb{D}} \mathcal{L}(g,d),\ \text{where} 
\end{equation}
\vspace{-0.05cm}
\begin{equation}
\mathcal{L}(g,d) = \mathbb{E}_{x\sim P_{data}(x)}[\phi (D_d(x))] + \mathbb{E}_{x\sim G_{g}(z)}[\phi(1-D_d(x))] \nonumber
\end{equation}

This is accomplished via a gradient-based learning process whereupon $D_d$ learns a binary classifier that is the best possible discriminator between real and fake
data.  Simultaneously, it encourages $G_g$ to approximate the latent data
distribution. In general, both networks are trained by applying back-propagation.

\subsection{Related work}
\label{sec:related-papers}

Mode collapse and vanishing gradients are the most frequent GAN training pathologies~\cite{arjovsky2017towards,arora2017gans}, leading to inconsistent results. 
Prior studies tried to mitigate degenerate GAN dynamics with new generator or discriminator objectives (loss functions)~\cite{arjovsky2017wasserstein,mao2017least,nguyen2017dual,zhao2016energy} and applying heuristics~\cite{Radford2015unsupervised,karras2017progressive}.


Others have integrated EC into GAN training. 
Evolutionary GAN (E-GAN) evolves a population of generators~\cite{wang2018evolutionary}. 
The mutation selects among three optimization objectives (loss functions) to update the weights of the generators, which are adversarially trained against a single discriminator. 
Multi-objective E-GAN (MO-EGAN) has been defined by reformulating E-GAN training as a multi-objective optimization problem by using Pareto dominance to select the best solutions in terms of diversity and quality~\cite{baioletti2020multi}. 
Two genetic algorithms (GAs) have been applied to learn mixtures of heterogeneous pre-trained generators to specifically deal with mode collapse~\cite{toutouh2020}. 
Finally, in~\cite{garciarena2018evolved}, a GA evolves a population of GANs (represented by the architectures of the generator and the discriminator and the training hyperparameters). 
The variation operators exchange the networks between the individuals and evolves the architectures and the hyperparameters. 
The fitness is computed after training the GAN encoded by the genotype.

Another line of research uses CEAs to train a population of generators against a population of discriminators. 
Coevolutionary GAN (COEGAN) combines neuroevolution with CEAs~\cite{costa2019coevolution}. 
Neuroevolution is used to evolve the main networks' parameters.
COEGAN applies an \textit{all-vs-best} Comp-CEA (with \textit{k-best} individuals) for the competitions to mitigate the computational cost of \textit{all-vs-all}.

CEAs showed similar pathologies as the ones reported in GAN training, such as \textit{focusing}, and \textit{lost of gradient}, 
which have been attributed to a lack of diversity~\cite{popovici2012coevolutionary}.   
Thus, spatially distributed populations have been demonstrated to be particularly effective at maintaining diversity, while reducing the computational cost from quadratic to linear form~\cite{williams2005}.
\Lipi 
locates the individuals of a population of GANs (pairs of generators and discriminators) in a 2D toroidal grid. 
A neighborhood is defined by the cell itself and its adjacent cells according to Von Neumann neighborhood. 
Coevolution proceeds at each cell with sub-populations drawn from the neighborhood. 
Gradient-based learning is used to update the weights of the networks while evolutionary selection and variation are used for hyperparameter learning~\cite{schmiedlechner2018towards,schmiedlechner2018lipizzaner}. 
After each training iteration, the (weights of the) \textit{best} generator and discriminator are kept while the other sub-population members are refreshed by new copies from the neighborhood. A cell's update of its GAN is effectively propagated to the adjacent cells in four directions (i.e., North, South, East, and West) once the neighbors of the cell refresh their sub-populations from neighborhood copies. 
Thus, each cell's sub-populations are updated with new fit individuals, moving them closer towards convergence, while fostering diversity. 
Another approach, Mustangs, combines \Lipi and E-GAN~\cite{toutouh2019}. Thus, the mutation operator randomly selects among a set of loss functions instead of applying always the same one in order to increase variability. 
Finally, taking advantage of the spatial grid of \Lipi, a \textit{data dieting} approach has been proposed~\cite{toutouh2020_dieting}. 
The main idea is to train each cell with different subsets of data to foster diversity among the cells and to reduce the training resource requirements. 

In this study, we propose \ring, spatial distributed GAN training that uses a ring topology instead of a 2D grid. We contrast \ring to \Lipi in the next section. 


%% file: ring-gan.tex
\section{Ring-Distributed GAN Training}
\label{sec:ring-topology}

This section describes the \ring CEA GAN training, which applies the same principles (definitions and methods) as \Lipi~\cite{schmiedlechner2018lipizzaner}.
We introduce both 2D grid and ring topologies applied by \Lipi and \ring, respectively.
We summarize the spatially distributed GAN training method. 
We present the main distinctive features between \Lipi and \ring.

\subsection{Spatially Structured Populations}
\label{sec:populations}

\sloppy
\ring and \Lipi use a population of generators \hbox{$\mathbf{g}=\{g_1, \dots,
g_Z\}$} and a population of discriminators \hbox{$\mathbf{d}=\{d_1, \dots,
d_Z\}$}, which are trained against each other (where $Z$ is the size of the population). 
A generator-discriminator pair named \textit{center} is placed in each cell, which belongs to a ring in the case of \ring or to a 2D toroidal grid in the case of \Lipi. 
According to the topology's neighborhood, sub-populations of networks, generators and discriminators, of size $s$ are formed.  

For the $k$-th neighborhood, we refer the \textit{center} generator by $\mathbf{g}^{k,1}\subset \mathbf{g}$, the set of generators in the neighborhood by $\mathbf{g}^{k,2}, \ldots, \mathbf{g}^{k,s}$, and 
the generators in this $k$-th sub-population by $\mathbf{g}^k = \cup^{s}_{i=1} \mathbf{g}^{k,i} \subseteq
\mathbf{g}$. The same is stated for discriminators $\mathbf{d}$.

\begin{figure}[!h]
\begin{minipage}[c]{0.99\linewidth}
         \centering
          \includegraphics[width=0.95\linewidth]{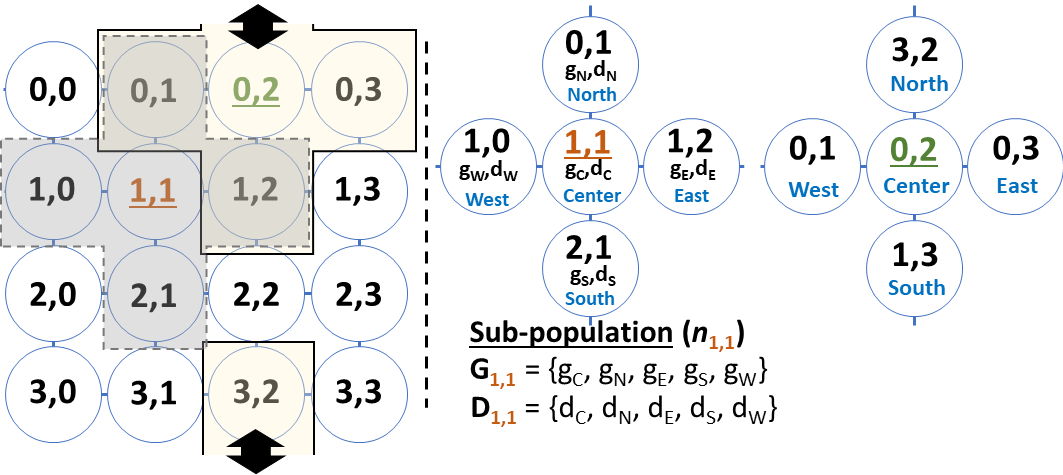}
         {a) Toroidal grid (neighborhood radius: 1) used by \Lipi.}
         \vspace{0.1cm}
\end{minipage}
\begin{minipage}[c]{0.99\linewidth}
         \centering
          \includegraphics[width=0.95\linewidth]{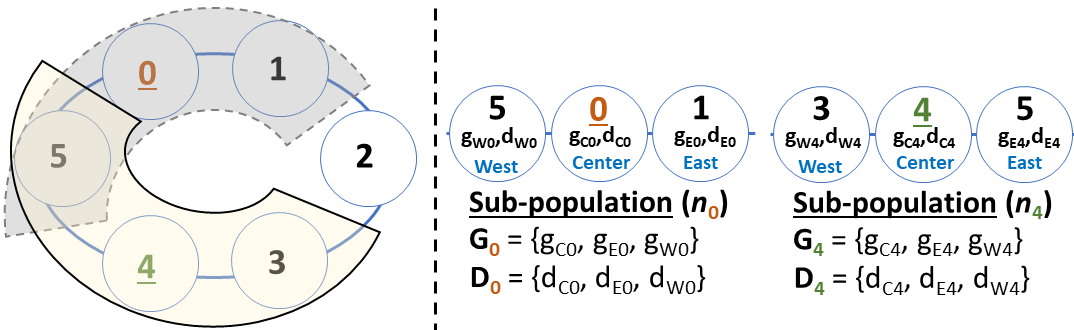}
         {b) Ring topology (neighborhood radius: 1). }
         \vspace{0.1cm}
\end{minipage}
\begin{minipage}[c]{0.99\linewidth}
         \centering
          \includegraphics[width=0.95\linewidth]{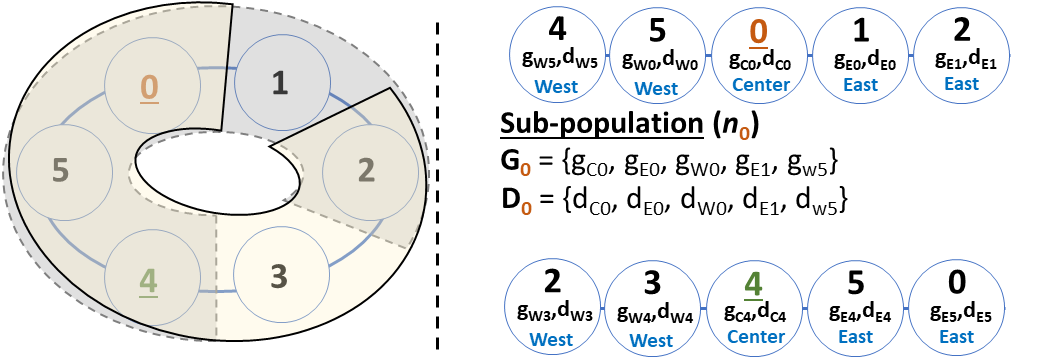}
         {c) Ring topology (neighborhood radius: 2). }
         \vspace{0.0cm}
\end{minipage}
\caption{Overlapping neighborhoods and sub-population definitions according topologies and neighborhood radius.}
  \label{fig:topology}
  \vspace{-.2cm}
\end{figure}

\Lipi uses Von Neumann neighborhoods with radius 1, which includes the cell itself and the ones in the adjacent cells to the North, South, East, and West~\cite{schmiedlechner2018lipizzaner}, i.e. $s$=5 (see Figure~\ref{fig:topology}.a).
This defines the migration policy (i.e., the directionality of signal propagation) through the cells in four directions.

Figure~\ref{fig:topology}.a shows an example of a 2D toroidal grid with $Z$=16 (4$\times$4 grid). The shaded areas illustrate the overlapping neighborhoods of the (1,1) and (0,2) cells, with dotted and solid outlines, respectively. The updates in the \textit{center} of (1,1) will be propagated to the (0,1), (2,1), (1,0), and (1,2) cells.   

In \ring, the cells are distributed in a one-dimensional grid of size $1\times Z$ and neighbors are sideways, e.g. left and/or right, i.e. an index position or more away. 
The best GAN (\textit{center}) after an evolutionary epoch at a cell is updated. Neighborhood cells retrieve this update when they refresh their sub-population membership at the end of their training epochs, effectively forming two non-ending pathways, around the ring, carrying signals in two directions. 
Figures~\ref{fig:topology}.b and~\ref{fig:topology}.c show populations of six individuals ($Z$=6) organized in a ring topology with neighborhood radius one ($r$=1) and two ($r$=2), respectively. 
The shaded areas illustrate the overlapping neighborhoods of the cells (0) and (4), with dotted and solid outlines, respectively. The updates in the \textit{center} of (0) will be propagated to the (5) and (1) for $r$=1 (Figure~\ref{fig:topology}.b) and to the (4), (5), (1), and (2) for $r$=2 (Figure~\ref{fig:topology}.c).

Thus the topologies of \Lipi and \ring influence communication (of the parameters of the best cell each epoch) and this, in turn, 
 affects how the populations converge.

\subsection{Spatially Distributed GAN Training}
\label{sec:populations}
\setlength{\intextsep}{0.6\baselineskip}

\alg~\ref{alg:lipizzaner} illustrates the main steps of the applied training algorithm. 
First, it starts the parallel execution of the training on each cell by initializing their
own learning hyperparameters (Line~\ref{alg:lipi-initialize}). 
Then, the training process consists of a loop with two main phases: first, \textit{migration}, in which the cells gather the GANs (neighbors) to build the sub-population ($n$), and second, \textit{train and evolve} when each cell updates the \textit{center} by applying the coevolutionary GANs training, (see \alg~\ref{alg:lipi-cea}). 
These steps are repeated $T$ (generations or training epochs) times.
After that, 
each cell learns an ensemble of generators by using an Evolutionary Strategy, \texttt{ES}-(1+1)~\cite[Algorithm 2.1]{loshchilov2013surrogate}, to compute the mixture weights $\omega$ to optimize the accuracy of the generative model returned $(n, \omega)^*$~\citep{schmiedlechner2018lipizzaner}.

\sloppy
For each training generation, the cells apply the CEA in \alg~\ref{alg:lipi-cea} in parallel. 
The CEA starts by selecting \textit{best} pair generator and discriminator (called $g_b$ and $d_b$) according to a tournament selection of size $\tau$. It applies an \textit{all-vs-all} strategy to evaluate all the GAN pairs in the sub-population according to a randomly chosen batch of data $Br$ (Lines~\ref{alg:lipi-random-batch} to~\ref{alg:lipi-select-center2}). 
Then, for each batch of data in the training dataset (Lines~\ref{alg:lipi-loop-batches} to~\ref{alg:lipi-update-d}), the learning rate $n_{\delta}$ is updated applying \textit{Gaussian Mutation}~\cite{schmiedlechner2018lipizzaner} and the offspring is created by training $g_b$ and $d_b$ against a randomly chosen discriminator and generator from the sub-population (i.e., applying gradient-based mutations).
Thus, the sub-populations are updated with the new individuals. 
Finally, a replacement procedure is applied to remove from the sub-populations the \textit{weakest} individuals and the \textit{center} is updated with the individuals with the best fitness (Lines~\ref{alg:lipi-update-subpops} to~\ref{alg:lipi-replace-center}). 
The fitness $\mathcal{L}_{g,d}$ of a given generator (discriminator) is evaluated according to the \textit{binary-cross-entropy} loss, where the model's objective is to minimize the Jensen-Shannon divergence between the \textit{real} and \textit{fake} data~\cite{goodfellow2014generative}.

\begin{algorithm}[h!]
	\small
	\caption{\Lipi main steps.\\
		\textbf{Input:} \hbox{T: Total generations,} \hbox{$E$: Grid cells,} \hbox{$s$: Neighborhood size,}
		\hbox{$\theta_{D}$: Training dataset,}
		\hbox{$\theta_{COEV}$: Parameters for \nameAlgLipiTraining,}
		\hbox{$\theta_{EA}$: Parameters for \texttt{MixtureEA}} \newline
		\textbf{Return:}
		~$n$: neighborhood, ~$\omega$: mixture weights
	}\label{alg:lipizzaner}
	\begin{algorithmic}[1]
		
		\ParFor{$c \in E$} \Comment{Asynchronous parallel execution of all cells in grid}
		\State $n, \omega \gets$ initializeCells($c, k, \theta_{D}$) \Comment{Initialization of cells}\label{alg:lipi-initialize}
		\For{generation} $\in [0,\dots, \text{T}]$ \Comment{Iterate over generations}
		 \label{alg:copy-neighboor}
         \State $n \gets$ copyNeighbours($c, k$) \Comment{Collect neighbor cells} \label{alg:copy-neighboor}
		\State $n \gets$ \nameAlgLipiTraining($n, \theta_{D}, \theta_{COEV}$) \Comment{Coevolve GANs}
		\EndFor
		\State $\omega \gets$ \texttt{MixtureEA}($\omega, n, \theta_{EA}$)\Comment{Build optimal ensemble}
		\EndParFor
		\State \Return $(n, \omega)^*$ \Comment{Cell with best generator mixture } 
	\end{algorithmic}
\end{algorithm}

\begin{algorithm}[h!]
	\small
	\caption{\nameAlgLipiTraining\newline
		\textbf{Input:}
		$n$: Cell neighborhood subpopulation,~$\theta_{D}$~: Training dataset,
		~$\tau$~: Tournament size,
		~$\beta$~: Mutation probability\newline
		\textbf{Return:}
		~$n$~: Cell neighborhood subpopulation trained
	}\label{alg:lipi-cea}
	\begin{algorithmic}[1]
	\State $Br \gets $ getRandomBatch($\theta_{D}$) \Comment{Random batch to evaluate GAN pairs } \label{alg:lipi-random-batch}	
	\For{$g, d \in \mathbf{g} \times \mathbf{d}$} \Comment{Evaluate all GAN pairs}
		\State $\mathcal{L}_{g,d} \gets$ evaluate($g, d, Br$) \Comment{ Evaluate GAN} \label{alg:fit-eval-1}
		\EndFor 
	  \State $g_b, d_b \gets$ select($n, \tau$) \Comment{Tournament selection} \label{alg:lipi-select-center2}
		
		\For{$B \in \theta_{D}$} \Comment{Loop over the batches in $\theta_{D}$}\label{alg:lipi-loop-batches}
		\State $n_{\delta} \gets$ mutateLearningRate($n_{\delta}, \beta$) \Comment{Update own $n_{\delta}$} 
		\State $d \gets$ getRandomOpponent($\mathbf{d}$) \Comment{Get random discriminator} \label{alg:lipi-rand-d}
		\State $g_b \gets$ updateNN($g_b, d, B$) \Comment{Update $g_b$ with gradient }
		\State $g \gets$ getRandomOpponent($\mathbf{g}$) \Comment{ Get uniform random generator} \label{alg:lipi-rand-g}
		\State $d_b \gets$ updateNN($d_b, g, B$) \Comment{Update $d_b$ with gradient } \label{alg:lipi-update-d}
		\EndFor
		\State $\mathbf{g}, \mathbf{d} \gets$ updatePopulations($\mathbf{g}$, $\mathbf{d}$, $g_b$, $d_b$) \Comment{Add $g_b$ and $d_b$} \label{alg:lipi-update-subpops}
		
		\For{$g, d \in \mathbf{g} \times \mathbf{d}$} \Comment{Evaluate all updated GAN pairs} \label{alg:fit-eval-1.5}
		\State $\mathcal{L}_{g,d} \gets$ evaluate($g, d, Br$) \Comment{ Evaluate GAN} \label{alg:fit-eval-2}
		\EndFor
		\State $n \gets$ replace($n, \mathbf{g}, \mathbf{d}$) \Comment{Replace the networks with worst loss}
        \State $n \gets$ setCenter($n$) \Comment{Best gen. and disc. are placed in the center} \label{alg:lipi-replace-center}
		\State \Return $n$
	\end{algorithmic}
\end{algorithm}

Here, we have summarized the spatially distributed method applied. 
More detailed definitions of this spatially distributed GAN training method can be found in~\cite{schmiedlechner2018towards,schmiedlechner2018lipizzaner}. 

\subsection{\ring vs. \Lipi}
\label{sec:ring-vs-lipi}

The population structured in a ring with $r$=2 provides a signal propagation similar to the \Lipi toroidal grid because both topologies and migration models allow the \textit{center} individual to reach four cells (two located in the West and two in the East in the case of this ring), see figures~\ref{fig:topology}.a and ~\ref{fig:topology}.b. Thus, they provide the same propagation speed and, as their sub-populations are of size $s$=5, the same selection pressure. 
The signal propagation of the ring with $r$=1 is slower because, after a given training step, the \textit{center} only reaches two cells (the one in the West and the one in the East), see Figure~\ref{fig:topology}.b. 
In this case, the sub-populations have three individuals, which reduces the diversity in the sub-population and accelerates convergence (it has 40\% fewer individuals than \Lipi sub-populations). 
Thus, \ring with $r$=1 reduces the propagation speed, while accelerating the population's convergence.

\ring with $r$=1 has two main advantages over \Lipi:
\textit{a)} it mitigates the overhead due to the communication is carried out only with two cells (instead of four) and the sub-populations are smaller, which reduces the number of operations for fitness evaluation and selection/replacement; 
and \textit{b)} like all \ring with any radius, it does not require to have a rectangular grid of cells, but $Z$ may be any natural number. 


Given the infeasibility of analyzing the change in selection and other algorithm elements, we proceed empirically. 

%% file: experimental-setup.tex
\vspace{-0.1cm}
\section{Experimental setup}
\label{sec:experimental-setup}

We evaluate different distributed CEA GAN training on image data sets:
the well known, MNIST~\cite{mnist} and CelebA~\cite{liu2015faceattributes}; 
and a dataset of chest X-ray images of patients with COVID-19~\cite{cohen2020covid}. 

In our experiments, we evaluate the following algorithms: \Lipi; 
two variations of \ring both with $r$=1, \ringone that performs the same training epochs than \Lipi and \ringtime that runs for the same computational cost (wall clock time) than \Lipi; and \ringfive which is the \ring with $r$=2 performing the same number of iterations than \Lipi. 
These represent a variety of topologies and migration models. 
\ringtime is analyzed to make a proper comparison between \ringone ($r$=1) and \Lipi taking into account the computational cost (time). 
Table~\ref{tab:train} summarizes the main characteristics of the \ring variations studied.

The parameters are set according to the authors of \Lipi~\cite{schmiedlechner2018lipizzaner,toutouh2019}.
Thus, all these CEAs apply a tournament selection of size two.
The main settings used for the experiments are summarized in Table~\ref{tab:net-arq}.

\begin{table}[!h]
	\vspace{-0.1cm}
	\centering
	\small
	\setlength{\tabcolsep}{6pt}
    \renewcommand{\arraystretch}{0.9} 
	\caption{\small \ring variations analyzed.}
	\label{tab:train}
	\vspace{-0.25cm}
	\begin{tabular}{lrrr}
	    \hline
	    \rowcolor{GAINSBORO}
   	    Method & $r$ & $s$ & Variation\\ 
  	    \ringone & 1 & 3 & Same number of iterations as 2D \Lipi \\
  	    \ringtime & 1 & 3 & Same computational time as 2D \Lipi \\
  	    \ringfive & 2 & 5 & Same number of iterations as 2D \Lipi \\
		\hline
	\end{tabular}
	\vspace{-0.1cm}
\end{table}

%
  
\begin{table}[!h]
	\centering
	\small
    \renewcommand{\arraystretch}{0.9} 
	\vspace{-0.1cm}
	\caption{\small Main GAN training parameters.}
	\label{tab:net-arq}
	\vspace{-0.25cm}
	\begin{tabular}{lrrr}
	    \hline
	    \rowcolor{GAINSBORO}
		Parameter & MNIST & CelebA & COVID-19  \\ 
		\rowcolor{ALICEBLUE}
		\multicolumn{4}{c}{{Network topology}} \\
		Network type  & MLP & DCGAN & DCGAN \\ 
		Input neurons & 64 & 100 & 100\\ 
		Number of hidden layers & 2 & 4 & 4 \\ 
		Neurons per hidden layer & 256 & 16,384 - 131,072  & 32,768 - 524,288 \\ 
		Output neurons  & 784 & 64$\times$64$\times$64 & 128$\times$128\\ 
		Activation function & $tanh$ & $tanh$ & $tanh$ \\ 	    \hline
	    \rowcolor{ALICEBLUE}
		\multicolumn{4}{c}{{Training settings}} \\ 
		Batch size  & 100  & 128 & 69\\ 
		Skip N disc. steps & 1 & - & -\\ 	    \hline
	    \rowcolor{ALICEBLUE}
		\multicolumn{4}{c}{{Learning rate mutation}} \\ 
		Optimizer & Adam  & Adam & Adam\\ 
		Initial learning rate & 0.0002  & 0.00005  & 0.0002\\ 
		Mutation rate & 0.0001 & 0.0001 & 0.0001  \\ 
		Mutation probability & 0.5 & 0.5& 0.5 \\ 
		\hline
	\end{tabular}
	\vspace{-0.1cm}
\end{table}

For MNIST experiments, the generators and the discriminators are multilayer perceptrons (MLP).  
The stop condition of each method is defined as follows:
\textit{a)} \ringone, \ringfive, and \Lipi perform 200~training epochs to evaluate the impact of the topology on the performance and computational time required; and
\textit{b)} \ringtime stops after running for the same time than \Lipi to compare the methods taking into account the same computational cost (time). 
The population sizes are 9, 16, and 25, which means \Lipi uses grid of sizes 3$\times$3, 4$\times$4, and 5$\times$5.  
Besides, 
we study the impact of the size of the populations on \ringone by training rings of size between 2 and 9.

For CelebA and COVID-19 experiments, 
deep convolutional GANs (DCGAN) are trained. DCGANs have much more parameters than MLP (see Table~\ref{tab:net-arq}). 
Here, we compare \ringone and \Lipi. They stop after performing 20~training epochs for CelebA and 1,000 for COVID-19 (because COVID-19 training dataset has much fewer samples). 
This will discern the differences in terms of performance and computational cost with more complex networks. 
In these cases, the population size is 9.

The experimental analysis is performed on a cloud computation platform
that provides 16~Intel Xeon cores 2.8GHz with 64 GB RAM and an NVIDIA Tesla P100 GPU with 16 GB RAM. 
We run multiple independent runs for each method. 

We have implemented all variations of the \ring by extending \Lipi
framework~\cite{schmiedlechner2018lipizzaner} using {Python}
and {Pytorch}~\cite{NEURIPS2019_9015}. 

%% file: experimental-analysis.tex
\section{Results and discussion}
\label{sec:experimental-analysis}

This section presents the results and the analyses of the presented GAN training methods. 
The first subsections evaluate them with the MNIST dataset. 
They are measured in terms of: the FID score, the diversity of the generated samples by evaluating the total variation distance (TVD)~\cite{li2017distributional}, and the diversity in the genome space (network parameters). 
Then, we analyze the CelebA and COVID-19 results with measurements of Inception Score (IS) and computational time.
The next subsection presents the results incrementally increasing the ring size of \ringone. 
Finally, we compare the computational times needed by \ringone and \Lipi.  

\subsection{Quality of the MNIST Generated Data}
\label{sec:fid}

Table~\ref{tab:fid-lipi-vs-ring} shows the best FID value from each of the 30 independent runs performed for each method. 
All the evaluated methods improve their performance when increasing the population size, while maintaining the same budget. 
This could be explained by diversity increasing during the training, as the populations get bigger.

\ringtime has the lowest (best) median and mean FID results. In turn, it returned the generative model that provided the best quality samples (minimum FID score) for all the evaluated population sizes. 
The second best results are provided by \ringone. 
This indicates that the methods with smaller sub-populations converge faster, even though the ring migration model ($r$=1) slows down the propagation of the best individuals. 

\begin{table}[t!]
\renewcommand{\arraystretch}{0.9} 
  \small
  \centering
  \caption{\small FID MNIST results in terms of best mean, standard deviation, median, interquartile range (Iqr), minimum, and maximum. Best values in bold. (Low FID indicates better samples)}
    \vspace{-0.3cm}
  \label{tab:fid-lipi-vs-ring}
  \begin{tabular}{lrrrrr}
      \hline
    \rowcolor{GAINSBORO}
    Method & Mean$\pm$Std & Median & Iqr & Min & Max \\
    \hline
\rowcolor{ALICEBLUE}
\multicolumn{6}{c}{Population size: 9} \\
\ringone & 35.45$\pm$6.74 & 35.02 & 7.66 & 22.16 & 51.09 \\ 
\ringtime & \textbf{33.42}$\pm$5.94 & \textbf{32.83} & 6.59 & \textbf{22.01} & \textbf{50.56} \\ 
\ringfive & 42.61$\pm$6.74 & 42.67 & 8.70 & 33.44 & 58.90 \\ 
\Lipi & 40.43$\pm$6.41 & 40.33 & 10.37 & 30.00 & 52.04 \\ 
\hline
\rowcolor{ALICEBLUE}
\multicolumn{6}{c}{Population size: 16} \\
\ringone & 29.33$\pm$4.56 & 28.70 & 5.93 & 22.66 & 40.12 \\ 
\ringtime & \textbf{27.66}$\pm$4.31 & \textbf{27.30} & 2.23 & 17.38 & \textbf{39.03} \\ 
\ringfive & 32.02$\pm$5.86 & 31.63 & 9.63 & 20.52 & 43.76 \\ 
\Lipi & 31.84$\pm$7.26 & 30.80 & 6.77 & 19.15 & 52.51 \\ 
\hline
\rowcolor{ALICEBLUE}
\multicolumn{6}{c}{Population size: 25} \\
\ringone & 26.47$\pm$3.88 & 26.19 & 5.94 & 18.41 & \textbf{34.55} \\ 
\ringtime & \textbf{25.12}$\pm$4.65 & \textbf{24.54} & 5.84 & \textbf{16.65} & 39.89 \\ 
\ringfive & 28.01$\pm$4.04 & 28.84 & 2.90 & 19.77 & 34.92 \\ 
\Lipi & 28.81$\pm$4.86 & 28.14 & 7.32 & 22.56 & 40.57 \\ 
\hline
  \end{tabular}
  \vspace{-.3cm}
\end{table}

Comparing \ringfive and \Lipi, they provide close results. Though they use different topologies, their signal propagation and selection operate equally.

\begin{figure*}[!h]
\centering
\begin{minipage}[c]{0.31\textwidth}
         \centering
         \includegraphics[width=\textwidth]{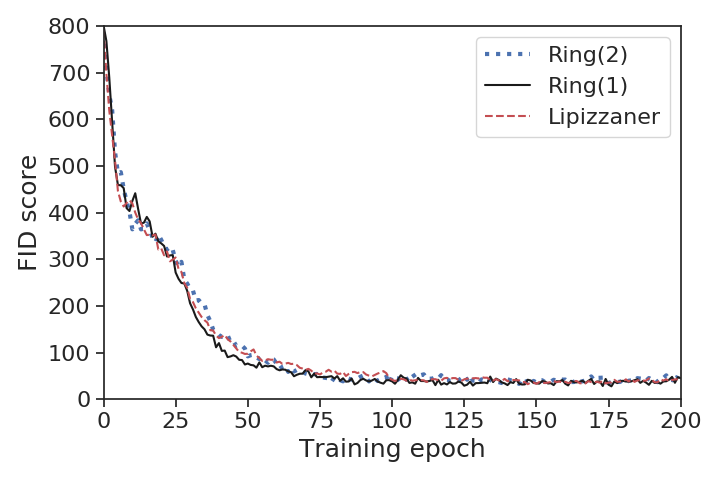}
         {a) Population size: 9}
\end{minipage}
\begin{minipage}[c]{0.31\textwidth}
         \centering
         \includegraphics[width=\textwidth]{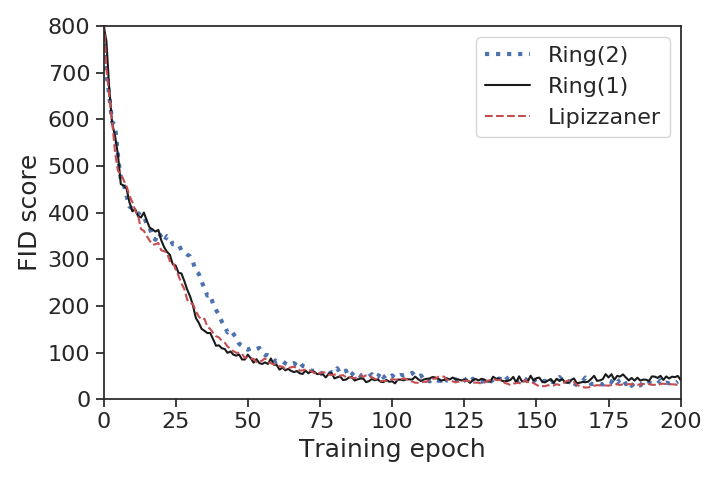}
         {b) Population size: 16}
\end{minipage}
\begin{minipage}[c]{0.31\textwidth}
         \centering
         \includegraphics[width=\textwidth]{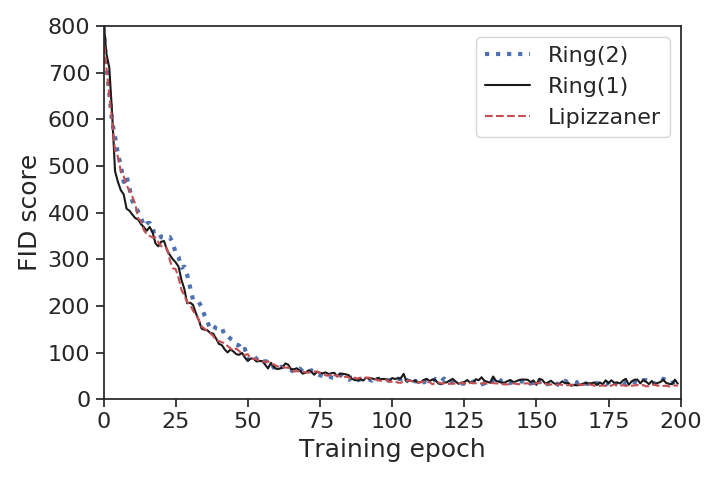}        
         {c) Population size: 25}
\end{minipage}
\vspace{-.3cm}
\caption{Median FID evolution.}
\vspace{-.1cm}
\label{fig:fid-evolution}
\vspace{-.cm}
\end{figure*}

As the results do not follow a normal distribution, we rank the studied methods using the Friedman rank statistical test and we apply Holm correction post-hoc analysis to asses the statistical significance. 
For all the population sizes, the Friedman ranks \ringtime, \ring, \Lipi, and \ringfive as the firsts, second, third, and fourth, respectively. 
However, the significance (p-value) varies from p-value=5.67$\times$10$^{-6}$ for population size 9 to p-value=1.13$\times$10$^{-2}$ (i.e., p-value$\ge$0.01) for population sizes 16 and 25. 
According to the Holm post-hoc analysis, \ringone and \ringtime are statistically better than \ringfive and \Lipi (which have no statistical differences between each other) for population size 9. 
For the other population sizes, \ringtime provides statistically better results than \ringfive and \Lipi and there is no significant difference among \ringone, \ringfive, and \Lipi. 

Table~\ref{tab:fid-lipi-vs-ring} shows \ringtime is better than \ringfive and \Lipi (lower FID is better). However, the difference between their FIDs decreases when population size increases. 
This indicates that migration modes that allow faster propagation take better advantage of bigger populations. 

Next, we evaluate the FID score throughout the GAN training process.
Figure~\ref{fig:fid-evolution} illustrates the changes of the median FID during the training process. \ringtime is not included because it operates the same as \ringone. 

According to Figure~\ref{fig:fid-evolution}, none of the evolutionary GAN training methods seem to have converged. 
Explaining this will be left to future work. 
The FID score almost behaves like a monotonically decreasing function with oscillations. 
The methods with larger sub-populations, i.e., \ringfive and \Lipi, show smaller oscillations which implies more robustness (less variance). 


Focusing on the methods with a ring topology, we clearly see the faster convergence when $r$=1. 
\ringone most of the time provides smaller FID values than \ringfive. 
The reduced sub-population ($s$=3) favors the best individual from the sub-population to be selected during the tournament (it increases the selection pressure). 

Figure~\ref{fig:fid-evolution} shows \ringone, \ringfive, and \Lipi converge to similar values. 
This is in accordance with the results in Table~\ref{tab:fid-lipi-vs-ring} that indicate these three methods provide comparable FID scores. 

\begin{figure}[!h]
\centering
\begin{minipage}[c]{0.23\textwidth}
         \centering
         \includegraphics[trim=0 8.5cm 0 0pt, clip, width=0.88\textwidth]{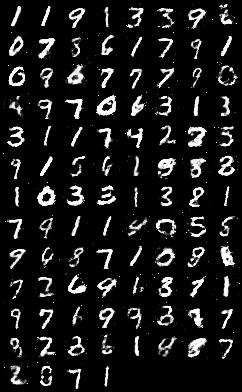}
         
         {a) \ringone MNIST samples}
\end{minipage}
\begin{minipage}[c]{0.23\textwidth}
         \centering
         \includegraphics[trim=0 8.5cm 0 0pt, clip, width=0.88\textwidth]{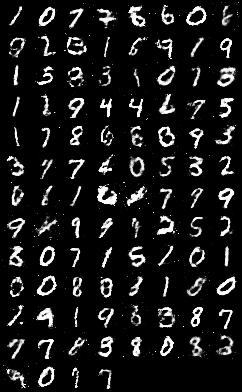}
         
         {b) \ringtime MNIST samples}
\end{minipage}

\vspace{0.2cm}
\begin{minipage}[c]{0.23\textwidth}
         \centering
         \includegraphics[trim=0 8.5cm 0 0pt, clip, width=0.88\textwidth]{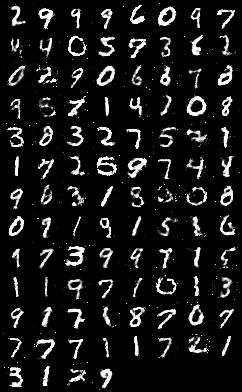}
         
         {c) \ringfive MNIST samples}
\end{minipage}
\begin{minipage}[c]{0.23\textwidth}
         \centering
         \includegraphics[trim=0 8.5cm 0 0pt, clip, width=0.88\textwidth]{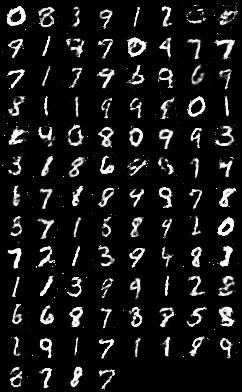}
         
         {d) \Lipi MNIST samples}
\end{minipage}
\caption{MNIST samples synthesized by generators trained using population size 16.}
\label{fig:mnist-samples}
\vspace{-0.2cm}
\end{figure}

Finally, Figure~\ref{fig:mnist-samples} illustrates some samples synthesized by generators trained using populations of size~16. As it can be seen, the four sets of samples show comparable quality.  

\subsection{Diversity of the Generated MNIST Outputs}
\label{sec:mnist-tvd}

This section evaluates the diversity of the samples generated by the best generative models obtained each run.  
Table~\ref{tab:tvd-lipi-vs-ring} reports the TVD for each method and population size.
Recall we prefer low TVDs (high diversity) to show the quality of the generative model.

\begin{table}[h!]
\renewcommand{\arraystretch}{0.9} 
  \small
  \centering
  \caption{TVD MNIST results in terms of best mean, standard deviation, Iqr, minimum, and maximum. Best values in bold. (Low TVD indicates more diverse samples).}
    \vspace{-0.2cm}
  \label{tab:tvd-lipi-vs-ring}
  \begin{tabular}{lrrrrr}
      \hline
    \rowcolor{GAINSBORO}
    Method & Mean$\pm$Std & Median & Iqr & Min & Max \\
    \hline
\rowcolor{ALICEBLUE}
\multicolumn{6}{c}{Population size: 9} \\
\ringone & \textbf{0.11}$\pm$0.02 &\textbf{ 0.11} & 0.03 & \textbf{0.07} & 0.16 \\ 
\ringtime & \textbf{0.11}$\pm$0.02 & \textbf{0.11} & 0.03 & \textbf{0.07} & \textbf{0.14} \\ 
\ringfive & 0.12$\pm$0.02 & 0.12 & 0.02 & 0.09 & 0.19 \\ 
\Lipi & 0.12$\pm$0.02 & 0.12 & 0.03 & 0.08 & 0.18 \\ 
\hline
\rowcolor{ALICEBLUE}
\multicolumn{6}{c}{Population size: 16} \\
\ringone & \textbf{0.09}$\pm$0.02 & \textbf{0.09} & 0.02 & \textbf{0.07} & 0.13 \\ 
\ringtime & 0.10$\pm$0.01 & \textbf{0.09} & 0.03 & 0.07 & \textbf{0.12} \\ 
\ringfive & 0.10$\pm$0.01 & 0.10 & 0.02 & \textbf{0.06} & 0.13 \\ 
\Lipi & 0.10$\pm$0.02 & 0.10 & 0.02 & 0.05 & 0.16 \\ 
\hline
\rowcolor{ALICEBLUE}
\multicolumn{6}{c}{Population size: 25} \\
\ringone &\textbf{ 0.09}$\pm$0.02 & \textbf{0.09} & 0.02 & \textbf{0.05} & 0.12 \\ 
\ringtime & \textbf{0.09}$\pm$0.02 & \textbf{0.09} & 0.02 & \textbf{0.05} & \textbf{0.11} \\ 
\ringfive & \textbf{0.09}$\pm$0.02 & \textbf{0.09} & 0.01 & 0.06 & 0.12 \\ 
\Lipi & \textbf{0.09}$\pm$0.02 & 0.10 & 0.02 & 0.06 & 0.16 \\ 
\hline
  \end{tabular}
\end{table}

The results in Table~\ref{tab:tvd-lipi-vs-ring} 
demonstrate that, as the population size increases, the resulting trained generative models are able to provide more diverse samples, 
that is, they have better coverage of the latent distribution. 
Thus, again, all the methods take advantage bigger populations. 

For population size 9, the mean and median TVD of \ringone and \ringtime are the lowest. 
According to Friedman ranking \ringtime, \ringone, \Lipi, and \ringfive are first, second, third, and fourth, respectively (p-value=0.0004). 
The Holm post-hoc correction confirms that there are no significant differences between \ringone and \ringtime and they are statistically more competitive than \ringfive and \Lipi (p-values$<$0.01).

For bigger populations, the statistical analyses report that there are no significant differences between the \ringone, \ringfive, and \Lipi, and \ringtime is statistically better than \ringfive and \Lipi (p-values$<$0.01).

With the support of the FID and TVD results, we can answer
\textbf{RQ1:} \textit{What is the effect on the generative model trained when changing the directionality of the signal propagation from four directions to two?} 
\textbf{Answer:} The impact on the results of performing migration to one or more neighbors is higher than the directionality itself. 
If we isolate the directionality, i.e., \ring vs. \Lipi, the main differences are revealed when $r$=1. 
Thus, \ringtime generators creates statistically better samples (FID and TVD results) than \Lipi. 
However, when $r$=2, there is no significant difference between \ringfive and \Lipi for all the evaluated population sizes. 
So, we observe that directionality is irrelevant. 

We can also answer \textbf{RQ2:} \textit{When the signal is propagated to only two directions, what is the impact of performing migration in one or more neighbors?} 
\textbf{Answer:} When comparing between \ring with $r$=1 and with $r$=2 using the same number of training epochs (i.e., \ringone and \ringfive), they perform the same although \ringone converges faster. 
The smallest sub-population of \ringone likely increases the selection pressure and the convergence speed of the sub-populations, despite slower signal propagation.

\begin{figure*}[!h]
\centering
\begin{minipage}[c]{0.2\textwidth}
         \centering
         \includegraphics[trim=0 20pt 0 10pt, clip, width=\textwidth]{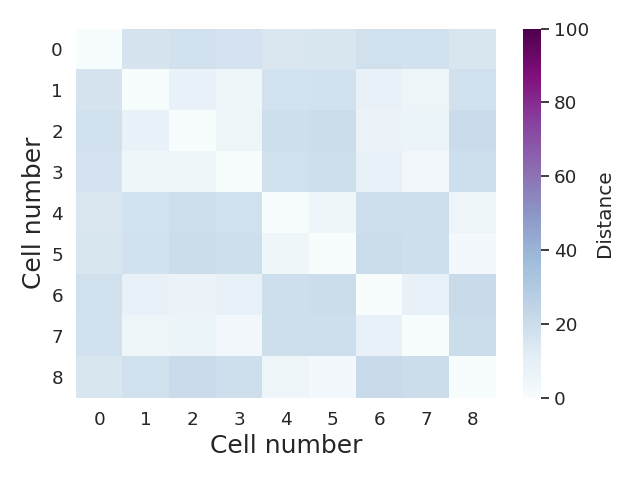}
         {a) \ringone(9)}
\end{minipage}
\hspace{0.01cm}
\begin{minipage}[c]{0.2\textwidth}
         \centering
         \includegraphics[trim=0 20pt 0 10pt, clip, width=\textwidth]{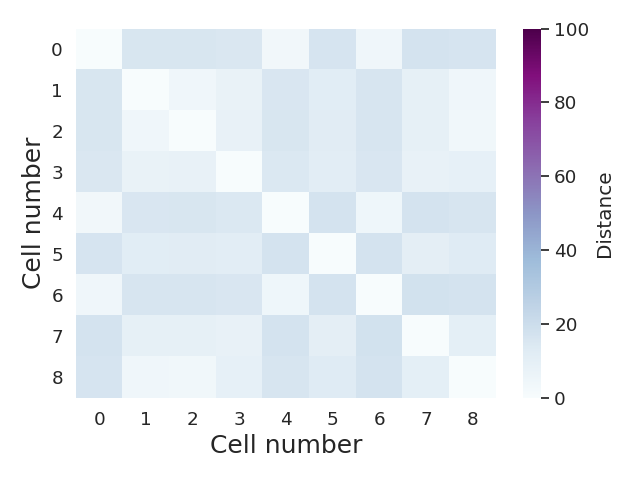}
         {b) \ringtime(9)}
\end{minipage}
\hspace{0.01cm}
\begin{minipage}[c]{0.2\textwidth}
         \centering
         \includegraphics[trim=0 20pt 0 10pt, clip, width=\textwidth]{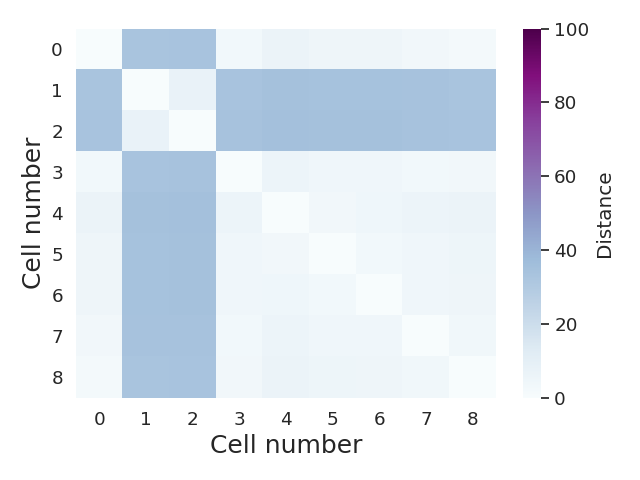}        
         {c) \ringfive(9)}
\end{minipage}
\hspace{0.01cm}
\begin{minipage}[c]{0.2\textwidth}
         \centering
         \includegraphics[trim=0 20pt 0 10pt, clip, width=\textwidth]{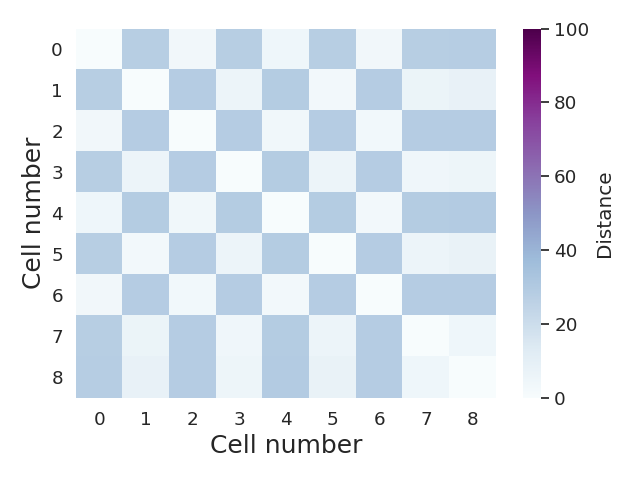}        
         {c) \Lipi(9)}
\end{minipage}
\vspace{0.1cm}

\begin{minipage}[c]{0.2\textwidth}
         \centering
         \includegraphics[trim=0 20pt 0 10pt, clip, width=\textwidth]{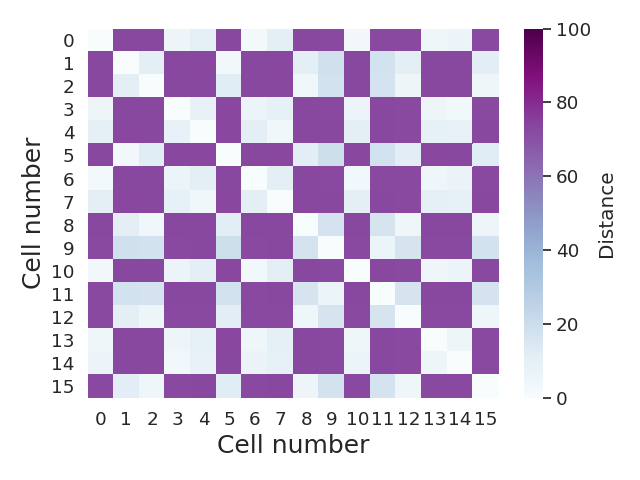}
         {a) \ringone(16)}
\end{minipage}
\begin{minipage}[c]{0.2\textwidth}
         \centering
         \includegraphics[trim=0 20pt 0 10pt, clip, width=\textwidth]{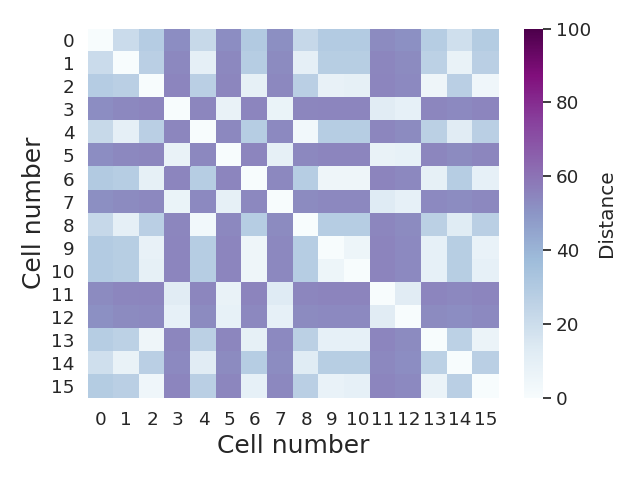}
         {b) \ringtime(16)}
\end{minipage}
\begin{minipage}[c]{0.2\textwidth}
         \centering
         \includegraphics[trim=0 20pt 0 10pt, clip, width=\textwidth]{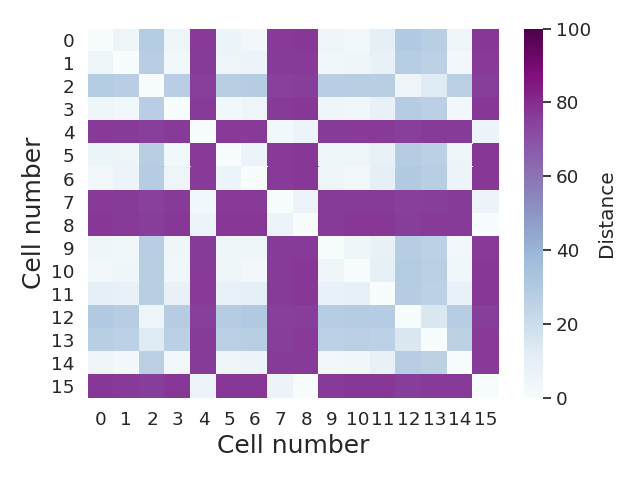}        
         {c) \ringfive(16)}
\end{minipage}
\begin{minipage}[c]{0.2\textwidth}
         \centering
         \includegraphics[trim=0 20pt 0 10pt, clip, width=\textwidth]{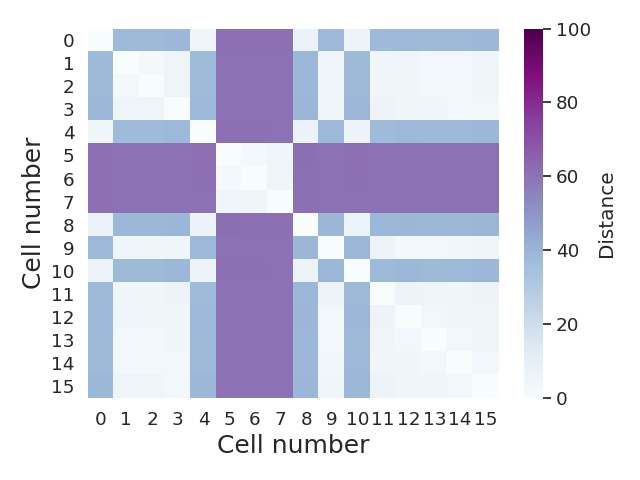}        
         {c) \Lipi(16)}
\end{minipage}
\caption{\small{Diversity in
            \textit{genome} space. Heatmap of $L_2$ distance between
            the generators for the population size 9 and 16 on MNIST at the end. X and Y axis show cell id. Dark indicates high distance. }}
  \label{fig:MNIST_weight_diversity_heat_map}
\end{figure*}

\subsection{Genome Diversity of MNIST}
\label{sec:genome-diversity} 
This section analyzes the diversity in the \textit{genome space} (i.e., the distance between the weights of the evolved networks). 
We evaluate the populations of size 9 and 16 to see how the migration model and sub-population size affect their diversity. 
$L_2$ distance between neural network parameters in a given population is used to measure diversity and Table~\ref{tab:distance-lipi-vs-ring} summarizes the results. 
Figure~\ref{fig:MNIST_weight_diversity_heat_map} presents the $L_2$ distances for the populations that represent the median $L_2$.

\begin{table}[h!]
\renewcommand{\arraystretch}{0.9} 
  \small
  \centering
  \caption{$L_2$ distance values between the weights of the networks in different cells on MNIST dataset. High values indicates more networks diversity. Highest value is in \textbf{bold}.}
    \vspace{-0.3cm}
  \label{tab:distance-lipi-vs-ring}
  \begin{tabular}{lrrrrr}
      \hline
    \rowcolor{GAINSBORO}
    Method & Mean$\pm$Std & Median & Iqr & Min & Max \\
    \hline
    \rowcolor{ALICEBLUE}
\multicolumn{6}{c}{Population size: 9} \\
\ringone & \textbf{16.92}$\pm$7.14 & 15.53 & 9.60 & \textbf{6.79} & \textbf{40.19} \\ 
\ringtime & 13.38$\pm$4.97 & 12.07 & 6.49 & 6.67 & 22.56 \\ 
\ringfive & 12.69$\pm$5.89 & 13.98 & 10.85 & 4.10 & 23.33 \\ 
\Lipi & 13.85$\pm$8.29 & \textbf{18.06} & 15.32 & 4.09 & 23.46 \\ 
\hline
    \rowcolor{ALICEBLUE}
\multicolumn{6}{c}{Population size: 16} \\
\ringone & \textbf{36.12}$\pm$18.72 & \textbf{34.21} & 30.32 & 13.44 & 70.28 \\ 
\ringtime & 34.88$\pm$13.33 & 32.62 & 10.30 & \textbf{17.75} & 66.18 \\ 
\ringfive & 32.44$\pm$21.10 & 21.97 & 25.90 & 11.31 & \textbf{86.03} \\ 
\Lipi & 30.99$\pm$24.28 & 28.09 & 55.82 & 2.72 & 61.34 \\ 
\hline
  \end{tabular}
\end{table}

Focusing on the \ring methods with $r$=1, the population diversity diminishes with more training, i.e., $L_2$ distances between the networks in \ringone are higher than in \ringtime for both population sizes, see Table~\ref{tab:distance-lipi-vs-ring} and Figure~\ref{fig:MNIST_weight_diversity_heat_map}. 
This shows that as the ring runs longer the genotypes are starting to converge. 

Taking into account the \ring methods with $r$=1 and $r$=2 that performed the same number of training epochs, i.e., \ringone and \ringfive, 
the first one shows higher $L_2$ distances (darker colors in Figure~\ref{fig:MNIST_weight_diversity_heat_map}). This confirms that the populations are more diverse when signal propagation is slower. 

Comparing \ringfive and \Lipi, that have the same sub-population size and migration to four neighbors, 
there is not a clear trend because \Lipi generated more diverse populations for population size 9 and \ringfive for population size 16. 
Therefore, we have not found a clear effect of changing the migration from four to two directions.

According to these results, we can answer the questions formulated in \textbf{RQ3}.
\textit{How does diversity change over time in a ring topology?} 
When \ringone performs more training  the diversity decreases.
\textit{How does diversity compare with a ring topology with different neighborhood radius?} 
As the radius increases, the diversity is lower because the propagation of the \textit{best} individuals is faster. 
\textit{How does diversity compare between ring topology and 2D grid, where both methods have the same sub-population size and neighborhood, but different signal directionality?}
We have not proven any clear impact on the diversity when changing the directionality for the methods that have the same sub-population size and neighborhood. 

\subsection{CelebA and COVID-19}
\label{sec:celeba-covid}

According to the empirical results, in general, \ringone and \Lipi compute generative models with comparable quality for MNIST (training MLP networks). Here, we study both methods training generative models for generating synthesized samples of CelebA and COVID-19 datasets using convolutional DNN, which are bigger networks (having more parameters).
Table~\ref{tab:ic-lipi-vs-ring-extra} shows the best IS value from each of the 10~independent runs for each method with population size 9.

In this case, it is more clear that both methods demonstrated the same performance. 
Focusing on CelebA dataset, \ringone computes higher (better) mean, minimum, and maximum IS, but \Lipi shows higher median. 
For the COVID-19 dataset, \Lipi shows better mean, median, and maximum IS values. 
The statistical analysis (ANOVA test) corroborates that there is not significant differences between both methods for both datasets (i.e., CelebA p-value=3.049 and COVID-19 p-value=0.731).

\begin{table}[h!]
\renewcommand{\arraystretch}{0.9} 
  \small
  \centering
  \caption{\small IS results in terms of best mean, standard deviation, median, Iqr, minimum, and maximum for CelebA and COVID-19 experiments. Best values in bold. (High IS indicates better samples)}
    \vspace{-0.3cm}
  \label{tab:ic-lipi-vs-ring-extra}
  \begin{tabular}{lrrrrr}
\hline
    \rowcolor{GAINSBORO}
    Method & Mean$\pm$Std & Median & Iqr & Min & Max \\
    \hline
\rowcolor{ALICEBLUE}
\multicolumn{6}{c}{Dataset: CelebA} \\
\ringone & \textbf{36.61}$\pm$1.95 & 35.94 & 2.31 & \textbf{34.40} & \textbf{40.40} \\ \Lipi & 36.30$\pm$1.77 &\textbf{ 36.34} & 2.86 & 32.95 & 39.02 \\ 
    \hline
\rowcolor{ALICEBLUE}
\multicolumn{6}{c}{Dataset: COVID-19} \\
\ringone & 1.79$\pm$0.12 & 1.79 & 0.20 & \textbf{1.63} & 2.00 \\ 
\Lipi & \textbf{1.86}$\pm$0.19 & \textbf{1.84} & 0.16 & 1.61 & \textbf{2.24} \\  
\hline
  \end{tabular}
\end{table}


%
%

\begin{figure}[!h]
\centering
\begin{minipage}[r]{0.22\textwidth}
         \centering
         \includegraphics[width=0.95\textwidth]{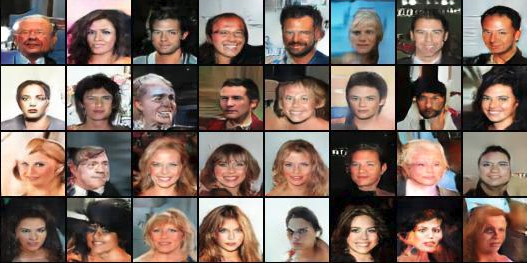}
         
         {a) \ringone - CelebA }
\end{minipage}
\hspace{0.05cm}
\begin{minipage}[l]{0.22\textwidth}
         \centering
         \includegraphics[width=0.95\textwidth]{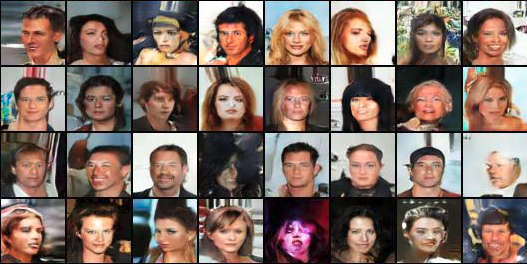}
         
         {b) \Lipi - CelebA }
\end{minipage}
\vspace{.2cm}

\centering
\begin{minipage}[r]{0.22\textwidth}
         \centering
         \includegraphics[trim=0 16pt 0 1pt, clip, width=0.95\textwidth]{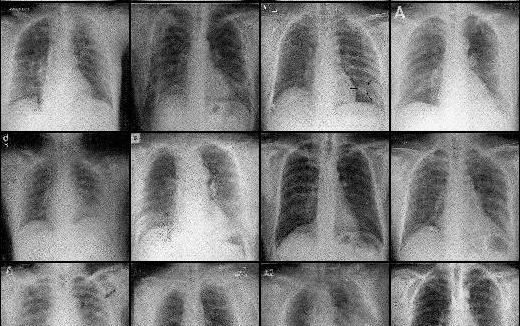}
         
         {c) \ringone - COVID-19 }
\end{minipage}
\hspace{0.05cm}
\begin{minipage}[l]{0.22\textwidth}
         \centering
         \includegraphics[trim=0 1pt 0 11pt, clip, width=0.95\textwidth]{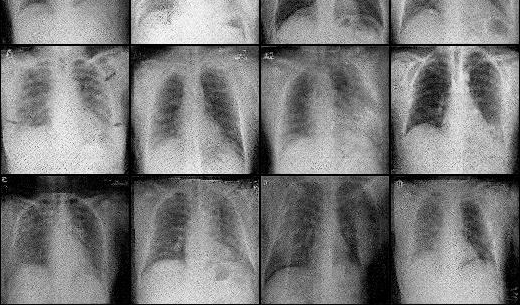}
         
         {d) \Lipi - COVID-19 }
\end{minipage}

\caption{Samples of CelebA and COVID-19 synthesized by generators trained using \ringone and \Lipi.}
\label{fig:covid-celeba-samples}
\vspace{-.3cm}
\end{figure}

Finally, Figure~\ref{fig:covid-celeba-samples} illustrates some samples of CelebA and COVID-19 synthesized by the generators trained using \ringone and \Lipi. As it can be seen, the samples show similar quality. 

Note that these results are in the line of the answers given to \textbf{RQ1} and \textbf{RQ2}. 
The change of directionality of the signal propagation and of the migration model allows the method to achieve the same results as \Lipi.

\subsection{\ring Scalability Analysis}
\label{sec:ring-scalability}

\begin{figure}[!h]
	\vspace{-0.1cm}
\centering
\includegraphics[trim=0 0pt 0 20pt, clip, width=0.4\textwidth]{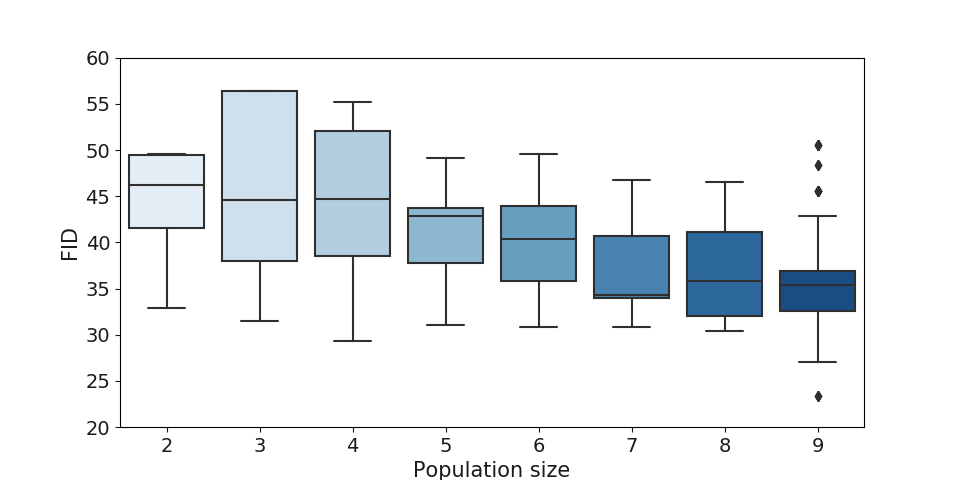}
\caption{FID MNIST results according to the population size.}
\label{fig:scalability}
	\vspace{-0.1cm}
\end{figure}

Here, we evaluate the results incrementally increasing the ring size of \ringone from 2 to 9 using MNIST dataset. 
Figure~\ref{fig:scalability} illustrates how increasing the population size of \ringone by only one individual improves the result (reduces the FID). However, \Lipi using, an a$\times$b 2D-grid would require adding (at least) $a$ or $b$ individuals.

%
\subsection{Computational Time Comparisons}
\label{sec:computational-performance}

We know that training with smaller sub-populations sizes takes shorter times because it performs fewer operations. 
As the reader may be curious about the time savings of using \ringone instead of \Lipi, 
we compare their computational times for the experiments performed. Notice that they perform the same number of training epochs (see Section~\ref{sec:experimental-setup}) and provide comparable quality results.  
Table~\ref{tab:time} summarizes the computational cost in wall clock time for all the methods, population sizes, and datasets. 
All the analyzed methods have been executed on a cloud architecture, which could generate some discrepancies.

\begin{table}[h!]
\renewcommand{\arraystretch}{0.9} 
  \small
  \centering
  \caption{Computational time cost in terms of mean and standard deviation (minutes).}
    \vspace{-0.3cm}
  \label{tab:time}
  \begin{tabular}{lrrr}
      \hline
    \rowcolor{GAINSBORO}
    \multicolumn{4}{c}{MNIST}\\
    \hline
\rowcolor{ALICEBLUE}
Population size & 9 & 16 & 25 \\
\ringone & \textbf{65.83}$\pm$1.02 & \textbf{72.95}$\pm$0.22 & \textbf{92.42}$\pm$1.82 \\
\Lipi & 87.86$\pm$1.28 & 91.19$\pm$1.80  & 105.54$\pm$1.98\\ 
\hline
    \rowcolor{GAINSBORO}
    \multicolumn{2}{c}{CelebA} & \multicolumn{2}{|c}{COVID-19}\\
    \hline
\rowcolor{ALICEBLUE}
Population size & 9 & \multicolumn{1}{|r}{Population size} & 9  \\
\ringone & \textbf{224.37}$\pm$2.36 & \multicolumn{1}{|l}{\ringone} & \textbf{118.87}$\pm$0.95 \\
\Lipi & 276.47$\pm$26.36 & \multicolumn{1}{|l}{\Lipi}  & 168.57$\pm$0.32\\ 
\hline
  \end{tabular}
\end{table}

As expected, \ringone requires shorter times.  
Comparing both methods on MNIST, \Lipi needed 33.46\%, 25.01\%, and 14.20\% longer times than \ringone for population sizes 9, 16, and 25, respectively. 
This indicates that the computation effort of using 40\% bigger sub-populations (5 instead of 3~individuals) affects less the running times as the population size increases. 


For CelebA and COVID-19 experiments,    
\ringone reduces the mean computational time by 23.22\% in CelebA experiments and by 41.18\% in COVID-19. 
The time saving is higher for COVID-19 mainly because the training methods performed more epochs (1,000) for this dataset than for CelebA (20~epochs). The sub-population size principally affects the required effort to perform the evaluation of the whole sub-population and the selection/replacement operation which are carried out for each training iteration. 
This explains why the time saving are higher for COVID-19 experiments.

%% file: conclusions.tex
\vspace{-0.1cm}
\section{Conclusions and future work}
\label{sec:conclusions}

The empirical analysis of different spatially distributed CEA GAN training methods shows that the use of a ring topology instead of a 2D grid does not lead to a loss of quality in the computed generative models, but it may improve them (it depends on the setup). 

\ringtime, which uses a ring topology with neighborhood radius $r$=1 and run for the same time than \Lipi, 
produced the best generative models. 
\ringone, \ringfive, and \Lipi, which were trained for the same training epochs, trained comparable generative models on the MNIST, CelebA, and COVID-19 datasets (similar FID and TVD for MNIST and IS for CelebA and COVID-19).

In terms of diversity, \ringone shows the most diverse populations which diminishes with more training. 
Focusing on ring topology, when the migration radius increases, i.e., \ringone ($r$=1) vs. \ringfive ($r$=2), the diversity decreases.
Finally, we have not found a marked difference on the diversity of the populations when changing the migration directionality , i.e., comparing between \ringfive and \Lipi.

\ringone, changing the signal propagation from four to two directions and using a migration of radius one, reduced the computational time cost of \Lipi by between 14.2\% and 41.2\%, while keeping comparable quality results. 

Future work will include the evaluation of \ringone on more datasets, bigger populations, and for longer training epochs. 
We will apply specific strategies to deal with small sub-populations (3~individuals per cell) to analyze the effect of reducing the high selection pressure.  
We will perform a convergence analysis to provide appropriate \Lipi and \ringone setups to address MNIST.  
Finally, we are exploring new techniques to evolve the network architectures during the CEA training.